\documentclass[conference,10pt]{IEEEtran}
\usepackage{pslatex}
\usepackage{graphicx}
\usepackage{pst-node}
\usepackage{graphics}
\usepackage[ngerman]{babel} 
\usepackage{eurosym}
\usepackage{amsmath}
\begin{document}

% paper title
\title{\huge CoZo+ - A Content Zoning Engine for Textual Documents}

% author names and affiliations
% use a multiple column layout for up to three different
% affiliations
\author{
\authorblockN{Cynthia Wagner, Christoph Schommer}\\
\authorblockA{{\small University of Luxembourg, Campus Kirchberg, CSC}\\
{\small 6, Rue Richard Coudenhove-Kalergi, L-1359 Luxembourg}\\
{\small Email: \{cynthia.wagner,christoph.schommer\}@uni.lu}\\
}\\
}

\maketitle
%\IEEEpeerreviewmaketitle
%\thispagestyle{empty}

\begin{abstract}
Content zoning can be understood as a segmentation of textual documents into zones. This is inspired by \cite{Teufel} who initially proposed an approach for the argumentative zoning of textual documents. With the prototypical Cozo+ engine, we focus on content zoning towards an automatic processing of textual streams while considering only the actors as the zones. We gain information that can be used to realize an automatic recognition of content for pre-defined actors. We understand Cozo+ as a necessary pre-step towards an automatic generation of summaries and to make intellectual ownership of documents detectable.
\end{abstract}

% =================================================================
\section{Introduction}\label{secIntroduction}

The basic idea of content zoning is to segment texts on the basis on pre-defined categories, called zones. A zone can be described as a set of sentences or paragraphs describing the same topic owning a zone name to reflect the topic. In a first approach to content zoning (\cite{BruWa07}), this has been applied to the analysis of spam emails - as means to separate spam from non-spam electronic mails. An important aspect in the zoning of texts is its structuring in itself - that has to be learned. Technically, a structuring is a composition of zones that are related to pre-defined actors, but that may form a fundament for continuing operations like the summarization of the story or the characterization of the authors. The described engine Cozo+ is a prototypical implementation. 

% =================================================================
\section{Cozo+ Architecture}\label{secCoZoArchitecture}
\begin{figure}[!htb]
\centering
\includegraphics[scale=0.4]{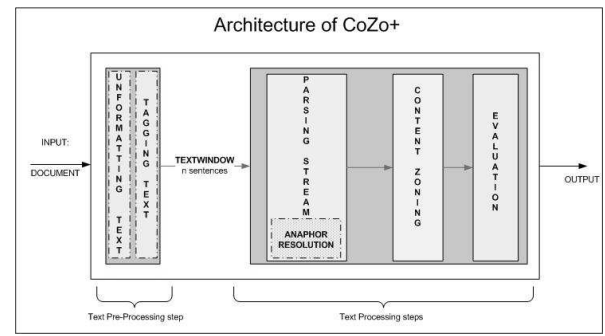}
\caption{CoZo+ Architecture}
\label{img:cozo}
\end{figure}

Cozo+ is a zoning engine and an extension to Cozo (\cite{BruWa07}). Cozo+ consists of two main modules, which is a pre-processing module and a text-stream processing module (see Figure \ref{img:cozo}). Generally, we have concerned with documents written in the english language only. 

In the text pre-processing module, these documents are read and sent to the application memory. We remove diverse formats of the document, for example paragraphs, line-breaks, etc. for having a continuous flow of text. Then, a Part-of-speech Tagger processes the prepared document to annotate the words with grammatical tags.  We consider a text stream as an endless flow of text from one point to another. For managing text streams, the user has to define a text window which indicates the text stream length for the streaming in unit sentences. This text window serves as basis for the text streams, where a text window of a size of n sentences fixed by the user, is read-in into the text processing module. 

The text stream processing module is composed of three sub-modules, namely a parser that is additionally featured with an anaphor resolution, the content zoning itself, and the evaluation module. The parsing of the stream is to determine the text into subject, verb, object relations in sentences. With these word relations, anaphors for the third person pronouns singular (he, she) and plural (they) can be resolved. The modified text stream is then passed to the content zoning module to extract content for user-defined actors/zones.

By applying grammatical search queries or rules, information about user-defined actors can be extracted into zones. After the zoning of a text stream, this text stream is lost. The last module in Cozo+ is the statistical evaluation of the zoned content for each zone, the mind-map structure generation with the statistical evaluation. The mind-map can be described as incremental and adaptive tree-like mind-map structure displaying the results of the zones. 

Zone variables can be described as parameters for analyzing \textit{zoned} content to gain further information about these zones. Diverse statistics about the size, a semantic and contextual analysis of the text content, the extraction of semantic or grammatical content of a zone, and much more can automatically be performed while finding relevant zones. In this respect, interesting zone variables are for example \textit{the most occurring word} or the \textit{most occurring sentence structure}.

An additional information in a zone is the gender (male, female) for the user-defined actor, used for pronoun resolution of main actors. An example for illustrating the effect of content zoning might be the following text having two actors \textit{Harry} and \textit{Hedwig}. Here, the raw text is read word by word, and the actor-based zones are assigned in case the actor name or a pronoun occurs or in case that the zone is still open (see \ref{secCoZoArchitecture}). If a pronoun - being an anaphor at this time - occurs, it must become solved (see \ref{aresolution}).

\begin{verbatim}
[Harry] Harry got up off the floor, stretched,
moved across to his desk. Harry neared the
bottom of the pile of newspapers. [/Harry] 
[Hedwig] Hedwig made no movement as she
began to flick through newspapers, throwing
them into the rubbish pile one by one. She
was asleep or else faking. She was angry about
the limited amount of time she was allowed out
of her cage at the moment. [Hedwig]
\end{verbatim}

As evaluation measurement a quality factor is introduced. The quality criterion is defined as a set of two parameters, representing the completeness and the correctness. They are called the \textit{matching} and the \textit{error rate}. The \textit{matching} is defined as a measure for \textbf{completeness} of the zoning, it is defined as to be \textit{complete} for human beings, which means that \textit{all} relevant sentences are zoned for a given actor, so that its resulting human matching is per se said to be exactly one.

\begin{equation}
Matching = \frac{ \mbox{Cozo+ zoned text}\; \bigcap_{}^{}{\mbox{Manually zoned text}} } {\mbox{Manually zoned text}}
\end{equation}

The \textit{error rate} is defined as a measure for \textbf{correctness} of a zoning. It is described as the ratio between wrongly zoned sentences for an actor (by Cozo+), the difference of all sentences of a given text, and the manually zoned sentences. The human zoning is considered to be 100\% \textit{correct}. A human zoning does not contain any erronous zoned sentences, so its resulting error-rate is per se said to be exactly null.

\begin{equation}
Error = \frac{ \mbox{wrongly zoned text (by Cozo+) } } {\mbox{complete text} \; - \; \mbox{Manually zoned text}}
\end{equation}

% =================================================================
\section{Implementation}\label{Implementation}
\begin{figure}[htbp]
\centering
\includegraphics[scale=0.2]{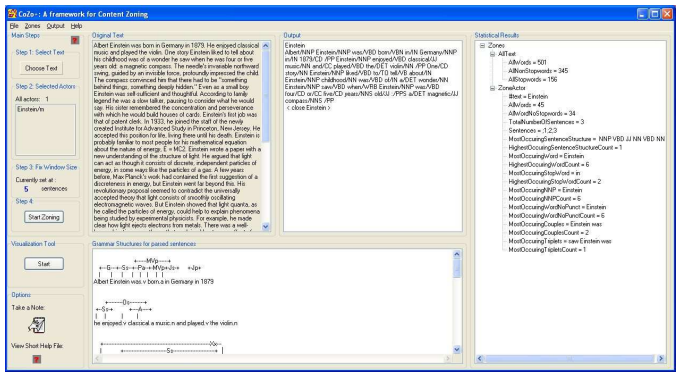}
\caption{Cozo+ Graphical User Interface with Setting and Result Windows}
\label{img:cozogui}
\end{figure}

% =================================================================
Cozo+ is implemented with a graphical user interface (see Figure \ref{img:cozogui}). This allows users to make their content zoning on text of different domains.

First, the user selects the text, the actors to be zoned, and finally the window size, which consists of a list of potential candidates for the pronoun resolution. The zones are then displayed in a separate window, as well as the parsed tree and diverse statistical values.

% =================================================================
\section{Results}\label{results}
% =================================================================
Cozo+ has been applied to different text domains, namely first online news articles, secondly, lifestyle articles about various topics, thirdly, biographies of famous inventors, fourth, fairy tales, fifth, theatre pieces, and, sixth, to scientific texts. The text size ranges between 300 up to 7000 words for a text. The shortest text category are online newspaper articles, the largest are chapters from books.

\label{aresolution} Anaphor resolution has been done in content zoning for incrementing the content zoning quality. A simple search query matching does not provide satisfying results. By resolving anaphors, more sentences can be zoned as this focuses on actors in texts. In this respect, results become improved through the resolution process of the third person singular (he/his/him/she/her) and plural (they/their). The success-rate for pronominal anaphors can be defined as the correctly solved anaphors for a given actor in Cozo+ concording with all manually counted anaphors for a given actor for an anaphoric category divided by all manually counted/solved anaphors for a given actor. The results of the success-rates for anaphor resolution in Cozo+ are illustrated in Table \ref{aresolution}.

\begin{table}[h]
\centering
\begin{tabular}{|l|l|}
\hline
\textbf{Pronominal anaphor} & \textbf{Average success-rate}\\
\hline
he/she & 72,2\% \\
\hline
his/him/her & 60,7\% \\ 
\hline
they/their & 65,3 \% \\
\hline
\end{tabular}
\label{aresolution}
\caption{Anaphor Resolution Accuracy Table}
\end{table}

\begin{table}[h]
\centering
\begin{tabular}{|l|l|l|}
\hline
\textbf{ } & \textbf{Human} & \textbf{Cozo+} \\
\hline
Actor & A.S. & A.S.\\
\hline
Counted sentences & 15 & 15 \\
\hline
Zoned Sentences & 9 & 7 \\
\hline
Erroneous zoned sentences & 0 & 0\\
\hline
Quality= & \{1 ; 0\} & \{0,78 ; 0\}\\
\{Matching ; Error-rate\} & & \\
\hline
\end{tabular}
\label{ManualvsContentZoning}
\caption{Manual versus content zoning results for \textit{Arnold Schwarzenegger} (A.S.)}
\end{table}

An observations is that the chosen anaphor method depends on the chosen text domains. In practical this can be explained as follows, easy texts of biographies achieve results, ranging between 80\% to 90 \% for the anaphor resolution of \textit{he}, \textit{she}, whereas complex texts, as chapters of \textit{Harry Potter} books, with a lot of speech passages, complex sentence structures, only achieve 45\% to 70 \% accuracy in the anaphor resolution for \textit{he},\textit{she}. Other observations are that once an anaphor is wrongly solved, this error can be propagated until an actor (re)-appears, but a worst case scenario is that the error can be propagated through an entire text window or even entire text stream. Another observation is the text window size. The anaphor resolution method used in Cozo+ does not store candidates from one text window to another, so by choosing a small text window, anaphors cannot be resolved if a new text window starts with an anaphor. These problems consequently decrease content zoning quality a lot.

\begin{figure}[h]
%a) \includegraphics[width=8.5cm]{cozo_out_news.jpg}\\ 
a) \includegraphics[scale=0.3]{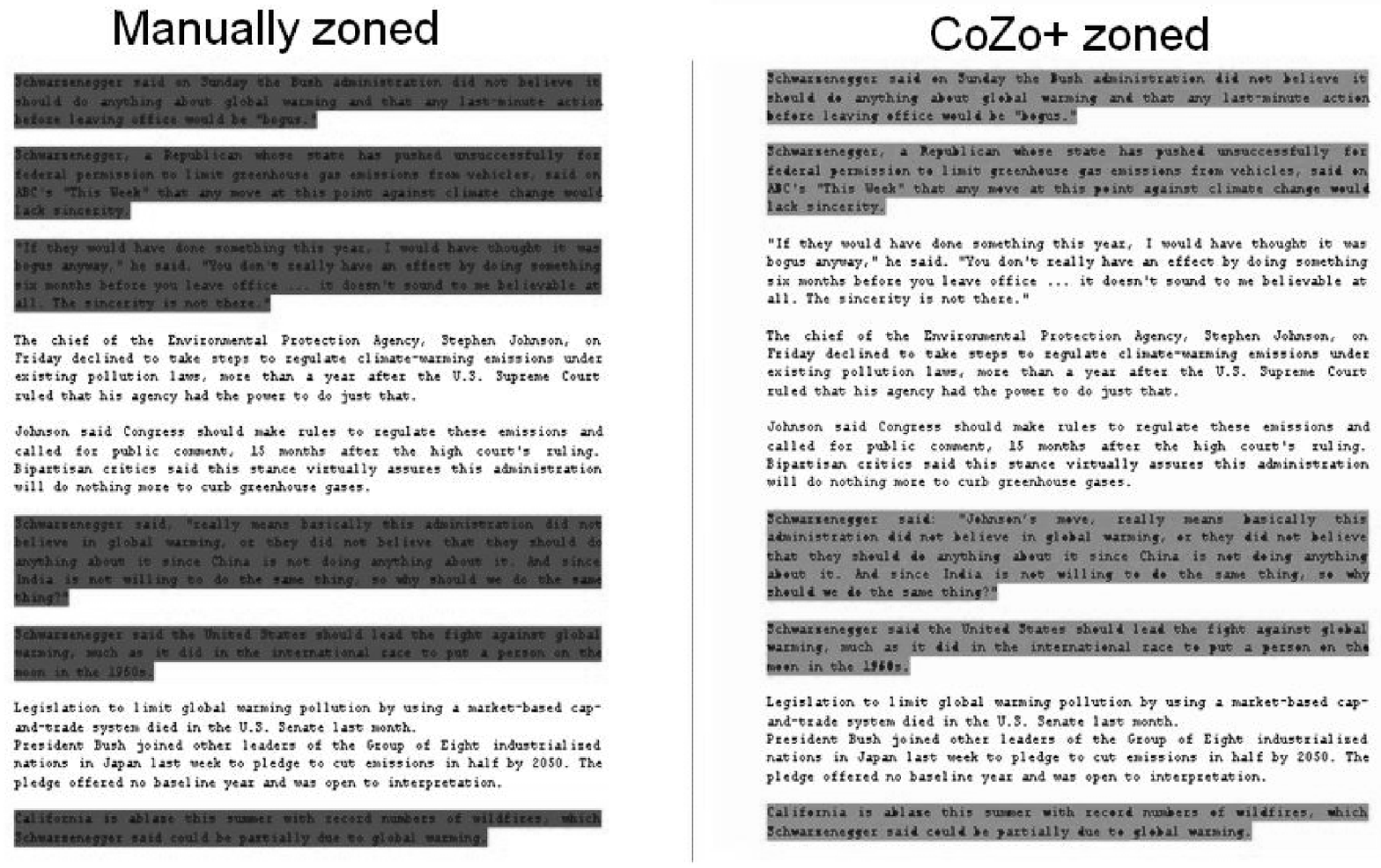}\\
%b) \includegraphics[width=8.1cm]{einstein_cozo.jpg}
b) \includegraphics[scale=0.2]{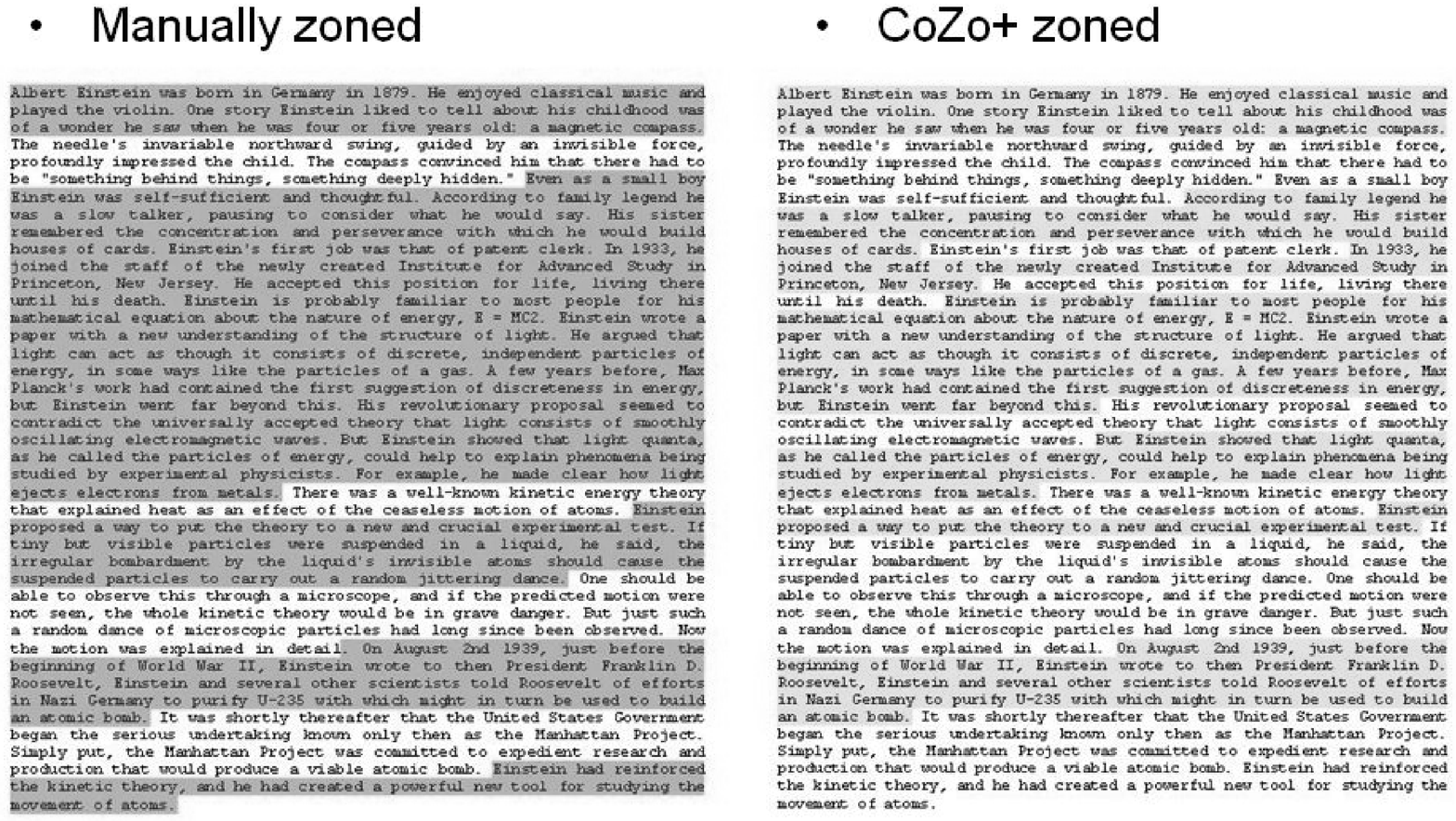}\\

\caption{Manual Zoning versus content zoning in a) news and b) biographies}
\label{img:humancozoex2}
\end{figure}

A first result obtained by using Cozo+ is the adaptive-incremental mind-map with its statistical and semantic results of the occurring actors in a text.  For evaluating Cozo+ results are compared to a human zoned text. Cozo+ is used to zone content from text on one side, on the other side a human being has zoned a text for some pre-defined actors. 

The results obtained for the matching and the error-rate in Cozo+ vary a lot, the results are probably domain-dependent. For illustrating, a Cozo+ content zoning versus a human zoning is shown in Figure \ref{ManualvsContentZoning}. The text is a middle-sized newspaper article. Figure \ref{img:humancozoex2} a) shows some extracted sentences for the pre-defined actor \textit{Schwarzenegger}. On the left side, the zoning is done manually, on the right side it is done by Cozo+. Table \ref{ManualvsContentZoning} regroups matching and error-rate. 

The values for human zoning matching, error-rate are said to be complete and correct as stated previously. For Cozo+, seven extracted sentences overlap with human extracted sentences; two sentences are missing in Cozo+ zoning. These sentences can be explained by the fact that Cozo+ does not have a module for recognizing speech without the indication of an actor. The error-rate is equal to null, because there are no wrongly zoned sentences for an actor when comparing manual, Cozo+ zoning. 

In a second example, a biography about \textit{Albert Einstein} is zoned once by human and afterwards by Cozo+. The pre-defined actor is \textit{Einstein}. Figure \ref{img:humancozoex2} illustrates the output for human and Cozo+.

\begin{table}[h]
\centering
\begin{tabular}{|l|l|l|}
\hline
\textbf{ } & \textbf{Human} & \textbf{Cozo+} \\
\hline
Actor & Einstein & Einstein\\
\hline
Counted sentences & 30 & 30 \\
\hline
Zoned Sentences & 20 & 15 \\
\hline
Erroneous zoned sentences & 0 & 0\\
\hline
Quality=\{Matching ; Error-rate\} & \{1 ; 0\} & \{0,75 ; 0\}\\
\hline
\end{tabular}
\caption{Human vs. content zoning results for \textit{Albert Einstein} (A.E.)}
\end{table}

\begin{table}[h]
\centering
\begin{tabular}{|l|l|}
\hline
\textbf{Text domain} & \textbf{Quality = }\\
  & \{ Avg. matching ; Avg. error-rate \}  \\
\hline
News & \{0,81 ; 0,02\} \\
\hline
Biographies & \{0,86 ; 0,0005\} \\
\hline
Lifestyle articles & \{0,86 ; 0,0067 \}\\
\hline
Scientific articles & \{0,563942 ; 0 \} \\
\hline
Fairy tales  & \{0,77 ; 0,002\}  \\
\hline
Theater pieces & \{0,36 ; 0,011\} \\
\hline
\end{tabular}
\caption{Total quality measures for Cozo+}
\label{Total}
\end{table}

In this example, sentences were not recognized by Cozo+. Possible explanations for this may be, the fixed text window size where actors are not stored from one text window to another, or the actor has not a subject or object position in the sentence. Table \ref{Total} regroups the evaluation of a human versus a Cozo+ zoning obtained for the different literature domains. The values for the quality are average measures from the qualities obtained from the different analyzed texts. 

In \ref{Total}, we observe that quality values vary a lot when considering the different literature domains. Biographies and lifestyle articles have high quality values due to the easy sentence structures with lot potential on actors. The same for biographies, where simple sentence structures and only few anaphors descibed the main actor. In scientific articles and theatre pieces less accurate quality results are achieved. In scientific articles commonly technology or research is discussed and not actors, so more accurate results could be obtained by using contextual zones and necessiting the \textit{it}-anaphor resolution. In theatre pieces, no classical sentence structures is followed, like subject-verb-object structures, but the main actor generally preceedes the sentence, which complicates the parsing task. 

The overall high zoning rates do not imply that the extracted text quantity must be high. The extracted quantity of text varies a lot, it is also depending on the pre-defined amount of actors. This can be explained as follows: the extracted quantity of text for the actor Benz in Karl Benz' biography is 80.95\% of the entire text, whereas the extracted text quantity for actor Harry in the \textit{Harry Potter, the deadly hallows, chapter 8} is only 11.3\% for the overall chapter. No direct relation between quality of zoned sentences for an actor and overall extracted text quantity of sentences for an actor can be observed. As a conclusion concerning the human performance versus the Cozo+ evaluation, Table \ref{compare} shows a regrouping of different quality influencing factors. 

\begin{table}[!h]
\centering
\begin{tabular}{|l|l|l|}
\hline
\textbf{Description} & \textbf{Yes} & \textbf{No} \\
\hline
Entire Text stream length &  & X \\
\hline
Text Domain & X &  \\
\hline
Text window size & X &  \\
\hline
Anaphor Resolution & X & \\
\hline
Complexity of text streams  & X &   \\
(for parsing) &   &   \\
\hline
Defined Actors &  & X \\
\hline
\end{tabular}
\caption{Quality influencing factors in Cozo+}
\label{compare}
\end{table}

% =================================================================
\section{Conclusions} \label{secConclusion}
% =================================================================
We have presented a engine for the zoning of content as it was initially introduced in \cite{BruWa07}. The engine consists of several modules that process incoming texts by firstly removing any formats, resolving pronouns to his corresponding actor, and finally outputs the texts including the zones. As it has been tested on various domains, the zoning results are quite promising and close to a manual zoning by humans. A next step will be the refinement of the zoning learning as the current zoning result is coarse-grained, keeping too much noise inside.

\section*{Acknowledgments}
The following work has been performed within a Master Thesis at the MINE Research Group of the ILIAS Laboratory, University of Luxembourg.

\end{document}